\documentclass[submission,copyright,creativecommons]{eptcs}
\providecommand{\event}{SOS 2007} 

\usepackage{iftex}

\usepackage{graphicx}
\usepackage{multirow}

\ifpdf
  \usepackage{underscore}         
  \usepackage[T1]{fontenc}        
\else
  \usepackage{breakurl}           
\fi

\title{Generating Causally Compliant Counterfactual Explanations using ASP\thanks{Authors supported by US NSF Grants IIS 1910131, US DoD, and industry grants.}}

\author{Sopam Dasgupta
\institute{Department of Computer Science\\
The University of Texas at Dallas\\
Texas, USA}
\email{sopam.dasgupta@utdallas.edu}\\
}

\def\titlerunning{Causality Compliant Counterfactuals}
\def\authorrunning{Sopam Dasgupta}
\begin{document}
\maketitle

\begin{abstract}
This research is focused on generating achievable counterfactual explanations. Given a negative outcome computed by a machine learning model or a decision system, the novel \textit{CoGS} approach generates (i) a counterfactual solution that represents a positive outcome and (ii) a path that will take us from the negative outcome to the positive one, where each node in the path represents a change in an attribute (feature) value. \textit{CoGS} computes paths that respect the causal constraints among features. Thus, the counterfactuals computed by \textit{CoGS} are realistic.  
\textit{CoGS} utilizes 
rule-based machine learning algorithms 
to model causal dependencies between features. 
The paper discusses the current status of the research and the preliminary results obtained.
\end{abstract}
\section{Introduction}
Predictive models used in automated decision-making processes (job-candidate filtering, loan approvals) often function as black boxes, making it difficult to understand their internal reasoning for decision-making. The decisions can have significant consequences, leading individuals to seek satisfactory explanations, especially for an unfavourable (negative) decision. Explaining these decisions presents a significant challenge. Additionally, users want to understand the changes necessary to flip a negative decision into a positive one.

Following Wachter et al.'s \cite{wachter} approach in this research, counterfactuals are employed to explain a machine learning model's reasoning behind a prediction. Counterfactuals help answer the question: ``What changes should be made to input attributes or features to flip a negative outcome to a positive one?" Counterfactuals also serve as a good explanation for a prediction. Wachter et al. \cite{wachter} use statistical techniques by examining the proximity of points in the N-dimensional feature space to find counterfactuals. This paper presents the \textit{Counterfactual Generation with s(CASP) (CoGS)} framework, which generates counterfactual explanations from \textit{rule-based machine learning (RBML)} algorithms such as FOLD-SE \cite{foldse}. \textit{CoGS} makes two advances compared to Wachter et al.'s work: (i) It computes counterfactuals using \textit{RBML} algorithms and ASP \cite{gelfond-kahl} rather than statistical techniques, and (ii) It considers causal dependencies among features when computing these counterfactuals. Another novelty of the \textit{CoGS} framework is that it further leverages the FOLD-SE algorithm \cite{foldse} to automatically discover potential dependencies between features that a user subsequently approves.


\textit{CoGS} models various scenarios (or worlds): the current \textit{initial state} $i$ represents a negative outcome, and the \textit{goal state} $g$ represents a positive outcome. A state is represented as a set of feature-value pairs. \textit{CoGS} finds a path from the \textit{initial state} $i$ to the \textit{goal state} $g$ by performing interventions (or transitions), where each intervention corresponds to changing a feature value while considering causal dependencies among features. These interventions ensure realistic and achievable changes that will take us from state $i$ to $g$. \textit{CoGS} relies on common-sense reasoning, implemented through answer set programming (ASP) \cite{gelfond-kahl}, explicitly using the goal-directed s(CASP) ASP system \cite{scasp-iclp2018}. The problem of finding these interventions can be viewed as a planning problem \cite{gelfond-kahl}, except that, unlike the planning problem, the moves (interventions) that take us from one state to another are not mutually independent.


\section{Background}\label{sec_background}
\smallskip\noindent\textbf{Counterfactual Reasoning:}\label{sec_cf}
Counterfactual reasoning is critical for explaining decisions in machine learning, offering insights on achieving desired outcomes by imagining plausible alternate scenarios. Wachter et al. \cite{wachter}
 advocated using counterfactual explanations to explain individual 
decisions, suggesting what changes could flip a negative outcome to a 
positive one. However, this approach often ignored causal dependencies, leading to unrealistic suggestions.
For a binary classifier given by $f:X \rightarrow \{0,1\}$, we define a set of counterfactual explanations $\hat{x}$ for a factual input $x \in X$ as $\textit{CF}_{f}(x)=\{\hat{x} \in X | f(x) \neq f(\hat{x})\}$. This set includes all inputs $\hat{x}$ leading to different predictions than the original input $x$ under $f$.

\smallskip\noindent\textbf{Causality Considerations:}
Causality relates to cause-effect relationship among predicates. $P$ is the cause of $Q$, if $(P \Rightarrow Q)$ $\wedge$ $(\neg P \Rightarrow \neg Q)$ \cite{SCM}. We say that $Q$ is causally dependent on $P$. 
Causality is crucial for generating realistic counterfactuals. For example, increasing the \textit{credit score} to be `high' while still being under increasing \textit{debt} obligations is unrealistic due to their causal link. Realistic counterfactuals must model these dependencies to ensure achievable changes.

\smallskip\noindent\textbf{ASP, s(CASP)}\label{dual_rules}
Answer Set Programming (ASP) is a paradigm for knowledge representation and reasoning \cite{cacm-asp,baral,gelfond-kahl}. ASP encodes feature knowledge, decision-making rules and causal rules, enabling the automatic generation of counterfactual explanations using this symbolic knowledge. \textbf{s(CASP)} is a goal-directed ASP system that executes answer set programs in a top-down manner without grounding \cite{scasp-iclp2018}. s(CASP) adopts \textit{program completion}, turning ``if'' rules $(P \Rightarrow Q)$ into ``if and only if'' rules ($(P \Rightarrow Q)$ $\wedge$ $(\neg P \Rightarrow \neg Q)$) which models causality.

\smallskip\noindent\textbf{FOLD-SE:}
FOLD-SE \cite{foldse}, is an efficient \textit{rule-based machine learning (RBML)} algorithm for classification tasks. It generates 
explainable models and learns causal rules from data. It maintains 
scalability and accuracy, making it a reliable component for the \textit{CoGS} framework, which leverages these rules for generating counterfactuals.

\smallskip\noindent\textbf{The Planning Problem:}
Planning involves finding a sequence of transitions from an initial state to a goal state while adhering to constraints. In ASP, this problem is encoded in a logic program with rules defining transitions and constraints restricting the allowed transitions \cite{gelfond-kahl}. Solutions are represented as a series of transitions through intermediate states. Each state is represented as a set of facts or logical predicates. Solving the planning problem involves searching for a path of transitions that meets the goal conditions within the constraints.
\textit{CoGS} can be thought of as a framework to find a plan---a series of interventions that change feature values---that will take us from the initial state to the final goal state. However, unlike the planning domain, the interventions (moves) are not independent of each other due to causal dependencies among features.  

\section{Research Goal}

\begin{figure}
    \centering    
    \includegraphics[width=15cm]{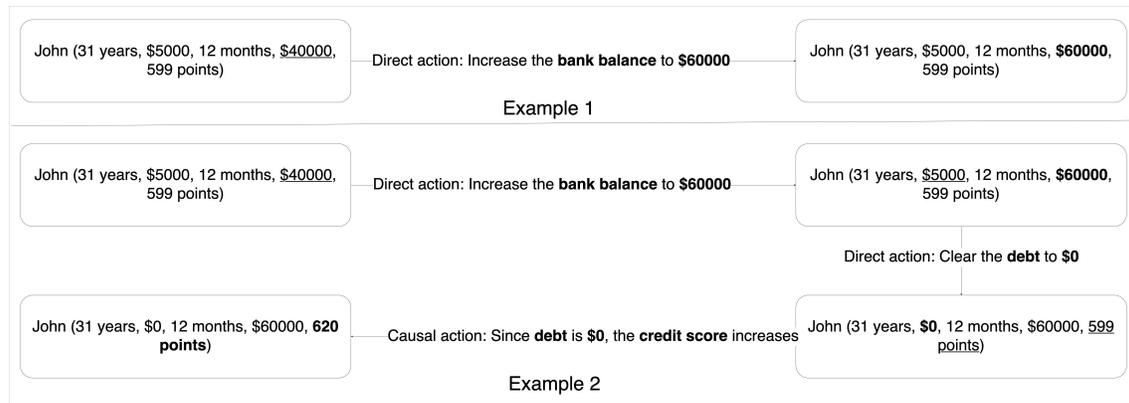}
    \caption{\textbf{Top:} Example 1 shows how John goes from being rejected for a loan to having his loan approved. Here the bank only considers the \textit{bank balance} for loan approval. John does a direct action to increase his \textit{bank balance} to $\$60000$. \textbf{Bottom:} Example 2 shows how John goes from being rejected for a loan to having his loan approved. Here the bank considers both \textit{bank balance} as well as \textit{credit score} for loan approval. While the \textit{bank balance} is directly altered by John, altering the \textit{credit score} requires John to directly alter his \textit{debt} obligations first. After clearing his \textit{debt}, the causal effect of having $\$0$ \textit{debt} increases John's \textit{credit score} to $620\ point$. This is the causal action}
    \label{fig_example_DC}
\end{figure}

This research aims to develop a framework that can encode feature knowledge, decision-making rules, and causal rules, enabling the automatic generation of counterfactual explanations using symbolic knowledge represented by an ASP program. The objective then is to use the ASP program to solve a version of the Planning Problem where the desired \textit{goal state} $g$ is our counterfactual state for solving the task at hand. This research focuses on generating counterfactuals for models that use decision rules (rule-based models). These rules are provided as explanations to justify decisions made by a governing authority, for example, a bank rejecting a loan due to a low \textit{bank balance} or a low \textit{credit score}. However, this can also be translated to statistical models by generating a rule-based approximation of these models.

Currently, most counterfactual-based approaches generate explanations without accurately accounting for the causal dependencies among features. These methods assume that the suggested changes will directly lead to a desired outcome, such as turning a negative decision into a positive one. As a result, these approaches are effective/practical only under two conditions: (1) when the features are independent or (2) when causal dependencies between features are irrelevant because only causally independent features are modified. For instance, consider Example 1 in Figure \ref{fig_example_DC}: John's loan application was rejected due to a \textit{bank balance} of $\leq\$60000$. The counterfactual solution suggests \textit{increasing} his \textit{bank balance} to $\$60000$. This recommendation is straightforward and achievable, as the \textit{bank balance} can be directly altered without affecting other features.

However, since most of these counterfactual-based approaches do not (accurately) model the causal dependencies between the features, changing certain features results in unintended changes to other features. Take Example 2 in Figure \ref{fig_example_DC}: John's loan application was rejected due to his poor ($\leq599\ points$) \textit{credit score}. The counterfactual solution tells him to \textit{increase} his \textit{credit score}. However, the \textit{credit score} is causally linked to the current \textit{debt} obligations and cannot be directly increased. Thus, the counterfactual solution may ultimately prevent the expected positive outcome from being achieved or even result in the generated counterfactual requiring a higher cost than initially assumed.

The proposed solution, \textit{CoGS}, would model the causal dependencies and provide a procedure/path informing in a step-by-step manner on what changes to make to achieve a counterfactual solution realistically. Example 2 assumes John has his loan application rejected due to his poor \textit{credit score}. The \textit{CoGS} solution tells him to \textit{clear} his \textit{debt} obligations. This increases John's \textit{credit score}, ultimately approving the loan. We approach this through the lens of the planning problem that provides us with a step-by-step path of the changes to make until we reach the goal state (counterfactual).


To summarize, the research goal is twofold: 1) Given the Decision Rules $D$ that give a negative outcome, we capture the causal dependencies $C$ amongst the features using user-defined rules or rules learnt using \textit{RBML} algorithms, and 2) Solve the planning problem where the \textit{goal state $g$} is defined as a state that is consistent with the causal rules $C$ and inconsistent with the decision rules $D$.

\section{Preliminary Results}\label{sec_experiments}

\begin{table}[t]
\centering
\fontsize{11}{11}\selectfont
\setlength{\tabcolsep}{2.5pt}
\begin{tabular}{@{}p{4cm} p{4.5cm} p{1cm} p{4.5cm} | p{1.2cm}@{}}
\hline
\\[-0.8em]
Features & Initial State & Action & Goal State & Time (ms) \\ 
\hline
\\[-0.8em]
Checking account status & $\geq\ 200$ & N/A & $\geq\ 200$ & \multirow{7}{*}{3236} \\ 
Credit history & no credits taken/all credits paid back duly & N/A & no credits taken/all credits paid back duly & \\ 
\textbf{Property} & real estate & \textbf{Direct} & car or other & \\ 
Duration months & 7 & N/A & 7 & \\ 
Credit amount & 500 & N/A & 500 & \\ 
Job & unemployed & N/A & unemployed & \\  
Present\hspace{0.1cm}Employment Since & unemployed/unskilled-non-resident & N/A & unemployed/unskilled-non-resident & \\ 
\hline
\end{tabular}
\caption{Transitions to goal states for the \textit{German} dataset: The value of \textit{Property} changes from \textit{real estate} to \textit{car or other}.}
\label{tbl_german}
\end{table}

\begin{table}[t]
\centering
\fontsize{11}{11}\selectfont
\setlength{\tabcolsep}{2.5pt}
\begin{tabular}{@{}l l l l l l | l@{}}
\hline
\\[-0.8em]
Features & Initial State & Action & Intermediate & Action & Goal State & Time (ms) \\ 
\hline
\\[-0.8em]
\textbf{Marital\_Status} & never\_married & N/A & never\_married & \textbf{Causal} & married\_civ\_spouse & \multirow{6}{*}{1126} \\
Capital Gain & \$6000 & N/A & N/A & N/A & $> 6849$ and $\leq 99999$ & \\ 
Education\_num & $7$ & N/A & N/A & N/A & $7$ & \\ 
\textbf{Relationship} & unmarried & \textbf{Direct} & husband & N/A & husband & \\ 
Sex & male & N/A & N/A & N/A & male & \\ 
Age & 28 & N/A & N/A & N/A & 28 & \\ 
\hline
\end{tabular}
\caption{Transitions to goal states for the \textit{Adult} dataset: The value of \textit{Relationship} changes from \textit{unmarried} to \textit{husband}. This has a causal effect of altering \textit{Marital Status} to \textit{married\_civ\_spouse}.}
\label{tbl_adult}
\end{table}

\begin{table}[t]
\centering
\fontsize{11}{11}\selectfont
\setlength{\tabcolsep}{2.5pt}
\begin{tabular}{@{}p{2cm} p{2.5cm} p{1cm} p{2.5cm} | p{2cm}@{}}
\hline
\\[-0.8em]
Features & Initial State & Action & Goal State & Time (ms) \\ 
\hline
\\[-0.8em]
persons & 4 & N/A & 4 & \multirow{4}{*}{1221} \\ 
\textbf{maint} & low & \textbf{Direct} & medium & \\ 
buying & medium & N/A & medium & \\ 
safety & medium & N/A & medium & \\ 
\hline
\end{tabular}
\caption{Transitions to goal states for the \textit{Car Evaluation} dataset: The value of \textit{maint} goes from \textit{low} to \textit{medium}.}
\label{tbl_car_eval}
\end{table}

We applied the \textit{CoGS} methodology to rules generated by the FOLD-SE algorithm (code on GitHub \cite{ref_supplement}). Our experiments use the German dataset \cite{german}, the Adult dataset \cite{adult}, and  the Car Evaluation dataset \cite{car}. These are popular datasets in the UCI Machine Learning repository \cite{ref_UCI}. The German dataset contains demographic data with labels for credit risk (`\textit{good}' or `\textit{bad}'), with records with the label `\textit{good}' vastly outnumbering those labelled `\textit{bad}'. The Adult dataset includes demographic information with labels indicating income (‘$=<\$50k/year$’ or ‘$>\$50k/year$’). The Car Evaluation dataset provides information on the acceptability of a used car being purchased. We relabelled the Car Evaluation dataset to \textit{`acceptable'} and \textit{`unacceptable'} to generate the counterfactuals.

For the (imbalanced) German dataset, the learned FOLD-SE rules determine a `\textit{good}' credit rating, with the undesired outcome being a `\textit{good}' rating since the aim is to identify criteria making someone a credit risk (`\textit{bad}' rating). Additionally, causal rules are also learnt using FOLD-SE and verified (for example, if the feature \textbf{`Job'} has the value  \textit{`unemployed'}, then the feature \textbf{`Present employment since'} should have the value  \textit{`unemployed/unskilled-non-resident'}). We learn the rules to verify these assumptions on cause-effect dependencies. 

\smallskip\noindent\textbf{Path to the Counterfactual:} By using these rules that identify individuals with a `\textit{good}' rating, we found a path to the counterfactuals, thereby depicting steps to fall from a \textit{`good'} to a \textit{`bad'} rating in Table \ref{tbl_german}. Similarly, we learn the causal rules and the rules for the undesired outcome for the \textit{Adult} dataset (undesired outcome: ‘$=<\$50k/year$’) as shown in Table \ref{tbl_adult}. For the \textit{Car Evaluation} dataset (undesired outcome: \textit{`unacceptable'}) shown in Table \ref{tbl_car_eval}, we only learn the rules for the undesired outcome as there are no causal dependencies (FOLD-SE did not generate any either). 
Tables \ref{tbl_german}, \ref{tbl_adult} and \ref{tbl_car_eval} show a path to each dataset's counterfactual goal state for a specific instance. Note that the execution time for finding the counterfactuals is also reported. While we have only shown specific paths in Tables \ref{tbl_german}, \ref{tbl_adult} and \ref{tbl_car_eval}, our \textit{CoGS} methodology can generate all possible paths from an original instance to a counterfactual.

\smallskip\noindent\textbf{Number of Counterfactual Sets:} Note that each path may represent a set of counterfactuals. This is because numerical features may range over an interval. Thus, \textit{CoGS} generates 240 sets of counterfactuals for the German dataset, 112 for the  Adult dataset, and 78 for the Car Evaluation dataset (Table \ref{tbl_counterfactual}).


\begin{table}[t]
\centering
\fontsize{11}{11}\selectfont
\setlength{\tabcolsep}{2.5pt}
\begin{tabular}{@{}l c | p{4cm}@{}}
\hline
\\[-0.8em]
Dataset & \# of Features Used & \# of Counterfactuals\\ 
\hline
\\[-0.8em]
Adult & 6 & 112\\ 
Cars & 4 & 78 \\
German & 7 & 240 \\
\hline
\end{tabular}
\caption{Table showing a Number of Counterfactuals produce by the \textit{is\_counterfactual} function given all possible states.}
\label{tbl_counterfactual}
\end{table}

\section{Related Work}
\vspace{-0.09 in}
Various methods for generating counterfactual explanations in machine learning have been proposed. Wachter et al. \cite{wachter} aimed to provide transparency in automated decision-making by suggesting changes individuals could make to achieve desired outcomes. However, they ignored causal dependencies, resulting in unrealistic suggestions. Utsun et al. \cite{ref_2_ustun} introduced algorithmic recourse, offering actionable paths to desired outcomes but assuming feature independence, which is often unrealistic. \textit{CoGS} rectifies this by incorporating causal dependencies. Karimi et al. \cite{alt_karimi} focused on feature immutability and diverse counterfactuals, ensuring features like gender or age are not altered and maintained model-agnosticism. 
 However, this method also assumes feature independence, limiting realism.
White et al. \cite{ref_clear} showed how counterfactuals can enhance model performance and explanation accuracy. Karimi et al. \cite{ref_4_karimi_2} further emphasized incorporating causal rules in counterfactual generation for realistic and achievable interventions. 
However, their method did not use the `if and only' property, which is vital in incorporating the effects of causal dependence. \textit{CoGS} rectified this by utilizing Answer Set Programming (ASP), which does not require grounding as it leverages s(CASP) to generate counterfactual explanations, providing a clear path from undesired to desired outcomes.

Bertossi \cite{ref_bertossi} utilizes Answer Set Programming (ASP) to generate \textit{causal explanations} by identifying \textit{minimal cardinality sets} using counterfactuals. These \textit{minimal cardinality sets} are used to compute scores to identify causal explanations. Unlike their work, \textit{CoGS} focuses on defining the causal dependencies amongst features and incorporating them into the framework. As a result of this \textit{CoGS} returns a series of steps to take to go from an original instance to a counterfactual instance which accounts for the causal impact of making interventions when going from one state to another.

The main contribution of this paper is the Counterfactual Generation with s(CASP) (\textit{CoGS}) framework for automatically generating counterfactuals while taking causal dependencies into account to flip a negative outcome to a positive one. \textit{CoGS} has the ability to find minimal paths by iteratively adjusting the path length.
This ensures that explanations are both minimal and causally consistent. \textit{CoGS} is flexible, generating counterfactuals irrespective of the underlying rule-based machine learning (\textit{RBML}) algorithm. The causal dependencies can be learned from data using any \textit{RBML} algorithm, such as FOLD-SE. The goal-directed s(CASP) ASP system plays a crucial role, as it allows us to compute a possible world in which a query {\tt Q} fails by finding the world in which the query {\tt not Q} succeeds. \textit{CoGS} advances the state of the art by combining counterfactual reasoning, causal modelling, and ASP-based planning, offering a robust framework for realistic and actionable counterfactual explanations. Our experimental results show that counterfactuals can be computed for complex models in a reasonable amount of time.

\section{Limitations and Planned Work}

One of the limitations of of \textit{CoGS} is its high computational time, which may lead to scalability issues. We are currently looking for ways to address this problem by replacing the multiple feature-independent values of a given feature with a single placeholder value. The plans for expanding on the work of \textit{CoGS} include:
\begin{itemize}
    \item Improving the execution time taken to generate counterfactual solutions as well as paths from the current outcome to the counterfactual instance. 
    \item Expanding \textit{CoGS} to generate counterfactuals for statistical machine learning methods: By running an \textit{RBML} algorithm on the predictions of the statistical model, a rule-based model approximation is generated. This approximation can then be used as the decision rules $D$ corresponding to the statistical model that is required by the \textit{CoGS} method.
    \item Improve the performance of machine learning systems: When machine learning models are trained on imbalanced datasets, the learned model often optimizes its performance on accurately predicting the majority class compared to the minority class. The plan is to generate counterfactual instances of the majority class, which will help us generate instances that belong to the minority class. 
    The expectation is that the machine learning model trained  on the modified training data will perform better with respect to both the majority and minority classes versus the original model trained on the original dataset (imbalanced).  

\end{itemize}

\section{Conclusion}

To conclude, this research is focused on automatically generating counterfactual solutions. This is accomplished by modelling causality and providing a path depicting the series of steps to be taken to achieve the counterfactual solution. 
This research proposes to do that by modelling the causal relationships that exist between features and the decision rules that led to the undesired negative outcome. Using these rules, a counterfactual solution is obtained. Finally, a version of the planning problem whose \textit{goal state} $g$ is the counterfactual solution, and the \textit{initial state} $i$ is the original negative outcome is solved. The generated plan corresponds to the path representing feature changes that take us from $i$ to $g$. These rules, as well as the modified planning problem, are modelled in s(CASP), a goal-directed answer set programming system.

\bibliographystyle{eptcs}
\bibliography{generic}
\end{document}